\documentclass{article}

    \usepackage[preprint]{neurips_data_2024}

\usepackage[utf8]{inputenc} 
\usepackage[T1]{fontenc}    
\usepackage{hyperref}       
\usepackage{url}            
\usepackage{booktabs}       
\usepackage{amsfonts}       
\usepackage{nicefrac}       
\usepackage{cancel}
\usepackage{microtype}      
\usepackage{xcolor} 
\usepackage{balance}
\usepackage{amsmath,amsfonts,amssymb }
\usepackage{algorithmic}
\usepackage{array}
\usepackage{algorithm}
\usepackage{algorithmic}
\usepackage{cleveref}
\usepackage{textcomp}
\usepackage{stfloats}
\usepackage{url}
\usepackage{verbatim}
\usepackage{graphicx}
\usepackage{subcaption}

\usepackage{amsmath,amsfonts,bm}

\def\eqref#1{equation~\ref{#1}}

\def\1{\bm{1}}

\newcommand{\tx}{\tilde{x}}
\newcommand{\ta}{\tilde{a}}
\newcommand{\tih}{\tilde{h}}
\newcommand{\talpha}{\tilde{\alpha}}

\newcommand{\tz}{\tilde{z}}
\def\ln{\text{ln}}

\DeclareMathAlphabet{\mathsfit}{\encodingdefault}{\sfdefault}{m}{sl}
\SetMathAlphabet{\mathsfit}{bold}{\encodingdefault}{\sfdefault}{bx}{n}

\def\gD{{\mathcal{D}}}

\def\gL{{\mathcal{L}}}

\def\gN{{\mathcal{N}}}

\def\gU{{\mathcal{U}}}

\def\sR{{\mathbb{R}}}

\newcommand{\E}{\mathbb{E}}

\newcommand{\sample}{\sim}

\title{Video Generation with Learned Action Prior
}

\author{%
  Meenakshi Sarkar\\
  Indian Institute of Science\\
  Bengaluru, India- 560054 \\
  \texttt{meenakshisar@iisc.ac.in} \\
  \And
  Devansh Bhardwaj \\
  Indian Institute of Technology \\
  Roorkee, India \\
  \texttt{d\_bhardwaj@ece.iitr.ac.in} \\
  \AND
  Debasish Ghose  \\
  Indian Institute of Science\\
  Bengaluru, India- 560054 \\
  \texttt{dghose@iisc.ac.in} \\
}

\begin{document}

\maketitle

\begin{abstract}
 Stochastic video generation is particularly challenging when the camera is mounted on a moving platform, as camera motion interacts with observed image pixels, creating complex spatio-temporal dynamics and making the problem partially observable. Existing methods typically address this by focusing on raw pixel-level image reconstruction without explicitly modelling camera motion dynamics. We propose a solution by considering camera motion or action as part of the observed image state, modelling both image and action within a multi-modal learning framework. We introduce three models: Video Generation with Learning Action Prior (VG-LeAP) treats the image-action pair as an augmented state generated from a single latent stochastic process and uses variational inference to learn the image-action latent prior; Causal-LeAP, which establishes a causal relationship between action and the observed image frame at time $t$, learning an action prior conditioned on the observed image states; and RAFI, which integrates the augmented image-action state concept into flow matching with diffusion generative processes, demonstrating that this action-conditioned image generation concept can be extended to other diffusion-based models. We emphasize the importance of multi-modal training in partially observable video generation problems through detailed empirical studies on our new video action dataset, RoAM.
\end{abstract}
\section{Introduction}
Video prediction is a valuable tool for extracting essential information about the environment and can be utilized by other learning frameworks such as motion planning algorithms \cite{Hafner2019}, and autonomous navigation and traffic management \cite{review_claussmann2020, long-prediction-traffic-cvpr18}. However, the complex interactions among different moving objects in a scene present significant challenges for long-term video prediction \cite{finn, finn2, mathieu, villegas, Gao_2019_ICCV, villegasNeurIPS2019,ebert17,sarkar2021}. Over the last decade, various works such as \cite{srivastava, oh, vondrick, finn, mathieu, villegas, WichersICML2018,video-pred-review-tpami2022,LiangICCV2017,ebert17} have tried to address this problem employing recurrent deep architectures and using concepts such as optical flow decomposition to adversarial training for generating high-quality output. 

\cite{Denton,BabaeizadehICLR2018,LeeICLR2018} have shown that video can be modelled as a latent stochastic process and variational inference can lead to reliable high-quality prediction on human action datasets such as KTH \cite{kth}, Human3.6M \cite{h36m_pami} and robotic datasets such as BAIR Robot Push \cite{ebert17}. However, these datasets only consider a static camera and do not capture the complexities arising from a moving camera.  \cite{villegasNeurIPS2019} showed that scaling deep stochastic variational architectures to a higher dimensional latent space can lead to better quality predictions in moving camera datasets such as KITTI. \cite{Gao_2019_ICCV} disentangled motion-specific flow propagation from motion-agnostic generation of pixel data for higher fidelity. With the success of transformer models in natural language processing, there is growing interest in designing efficient visual transformers \cite{ViT2021} for video prediction tasks \cite{video_transformer2022, Gao_2022_CVPR_ViT}. Recent works in diffusion-based video generation and prediction models \cite{ho2022diffusion,vidmAAAI2023,river2023ICCV,flexibleDiffusionNeurips2022,tmlr_hoppe22} have shown great promise in generating and predicting long-term high-fidelity videos.

In scenarios characterized by partial observability, especially when a camera is mounted on a moving platform, the captured image frames are influenced not only by the inherent scene dynamics but also by the actions of the mobile platform itself. This prevalent situation is particularly relevant in domains such as autonomous vehicles and mobile robotics. Numerous datasets, including KITTI \cite{kitti}, KITTI-360 \cite{KITTI-360}, A2D2 \cite{A2D2}, and the pedestrian video dataset by Caltech \cite{CalTech-pedestrian}, have already illustrated the significance of this scenario in the context of autonomous cars. More recently, Sarkar et al. \cite{acpnet2023} have underscored the importance of modelling partially observable video data within the domain of indoor robotics, as demonstrated by their RoAM dataset. This further emphasizes the growing relevance of addressing partial observability in various robotic applications. However most of the past works like \cite{sarkar2021,villegasNeurIPS2019,Gao_2019_ICCV,GaoCVPR2020} did not explicitly condition the generated future frames on the actions or movement of the recording camera even though they all highlighted the increased level of complexities arising from the interactions between the movement of the camera sensor and recorded image frames. This limitation can also be attributed, in part, to the absence of synchronized control action or actuator data in partially observable datasets like KITTI \cite{kitti} and A2D2 \cite{A2D2}. These datasets do not provide the concurrent recording of steering actions taken by autonomous cars while capturing video data.

However, the recently introduced  Robot Autonomous Motion or RoAM \cite{acpnet2023} dataset includes timestamped and synchronized control action data of the recording robotic agent along with stereo image pairs when the autonomous robot explored different indoor spaces capturing different human actions. This dataset presents the opportunity for vision scientists to design generative models that can be explicitly conditioned and model the on the robot action data in the context of video generation. Modelling this dynamic interaction is crucial for creating more accurate and realistic video predictions, especially in applications such as autonomous driving and robotic navigation.

In this paper, we propose a mathematical framework where the camera motion is considered as an extended part of the image state and we design generative models that can approximate the latent stochastic process with a learned image-action prior. We present 3 distinct models: (i) SVG-LeAP, (ii) Causal-LeAP and (iii) RAFI video diffusion model. SVG-LeAP and Causal-LeAP are both variational generative frameworks, with the key distinction being that Causal-LeAP considers a causal relationship between the action taken at time $t$, $a_t$, and the image observed, $x_t$. SVG-LeAP is based on the SVG-lp \cite{Denton} model with the modification of considering the image-action pair as the joint observed state of the system instead of just the image. With RAFI we extend the concept of a joint image-action system state to flow matching in diffusion generative processes. RAFI is built on the sparsely conditioned flow matching model of RIVER \cite{river2023ICCV}.
\section{Prior works}\label{sec:model}
Let us assume that $x_t$ is the current image frame from a sequence of frames $x_{1:T-1}$ from video data of dimension $d=[i_h\times i_w\times 3]$. Over the past decade, numerous mathematical frameworks have been proposed to model the generation process of $x_t$. In their seminal work, \cite{Denton} introduced the stochastic learned prior model (SVG-lp), which has gained widespread acceptance and application within the computer vision community. This framework posits that a sequence of image frames from a video is generated from a latent Gaussian distribution. The latent distribution is learned through a variational training and inference paradigm using a set of observed image sequences. Mathematically here the current image frame is predicted as $\tx_t$ conditioned on the past observed frames $x_{1:t-1}$ and a latent variable $z_t$. Given that at the time of prediction $p(z_t)$ is unknown, it is learnt with a posterior distribution $p_\theta(z_t|x_{1:t})= \gN(\mu_\theta(x_{1:t}),\sigma(x_{1:t}))$ approximated by a recurrent network parameterised by $\theta$. The sampled variable $z_t$ is then used to generate the current image frame $x_t$ conditioned on the past observed frames $x_{1:t-1}$. Denton proposed two methods for learning $p_\theta(z_t|x_{1:t})$: (i) with a fixed Gaussian prior and (ii) with a companion prior model $p_\phi(z_t|x_{1:t-1})$ and minimising the KL divergence loss between the two. 

This learned prior model has subsequently been utilized in various video generation models, such as those by \cite{villegasNeurIPS2019,chatterjee2021ICCV} in recent years. While effective, this model does not explicitly address the integration of camera motion or other modalities of the video data generation process into the architecture. Camera motion plays a crucial role in the video generation process, especially when the camera is moving or mounted on a moving platform like a car or a robot.   \cite{villegasNeurIPS2019} showed that with a significantly larger parametric space, SVG-lp can effectively generate and predict future image frames when tested on partially observable video datasets like KITTI, where the camera is mounted on a car. However, recent works, such as those by \cite{acpnet2023}, have demonstrated that long-term video prediction processes can be enhanced by explicitly conditioning the predicted frames on the motion of the camera.

Recently, diffusion models \cite{ho2022diffusion,river2023ICCV,MCVD2022neurips,song_iclr2021,sdvf_wacv2020,dmvp_tmlr2022} have garnered attention from the computer vision community due to their capacity to generate and forecast high-fidelity video sequences. Rooted in the concepts of diffusion processes \cite{ho2022diffusion}, these models iteratively refine noisy data to produce high-quality image frames.

\section{Action conditioned video generation}\label{sec:acvg}
We introduce three distinct action-conditioned video generation models. Our first two Learned Action Prior or LeAP models are stochastic video generation frameworks in which the action or camera movement is learned through a latent Gaussian distribution. However, the methods by which these action priors are learned differ significantly from each other, based on distinct sets of assumptions regarding the interaction between action and video. With the third model, we introduce RAFI, our Random Action-Frame Conditioned Flow Integrating video generation model, based on the RIVER diffusion framework by \cite{river2023ICCV}. RAFI showcases how camera actions can be seamlessly integrated into Flow Matching \cite{flowMatching2023ICLR} and the diffusion process to enhance video prediction quality.

In this paper, we denote the action of the robot or the platform on which the camera is mounted at timestep $t$ by $a_t\in \sR^{n}$, where $n$ is the dimension of the action or actuation space of the robot/platform. We also assume actions are normalised meaning $a_t \in [0,1]$.
\begin{figure}
    \centering
    \begin{subfigure}[t]{0.30\textwidth}
        \includegraphics[width=0.8\linewidth]{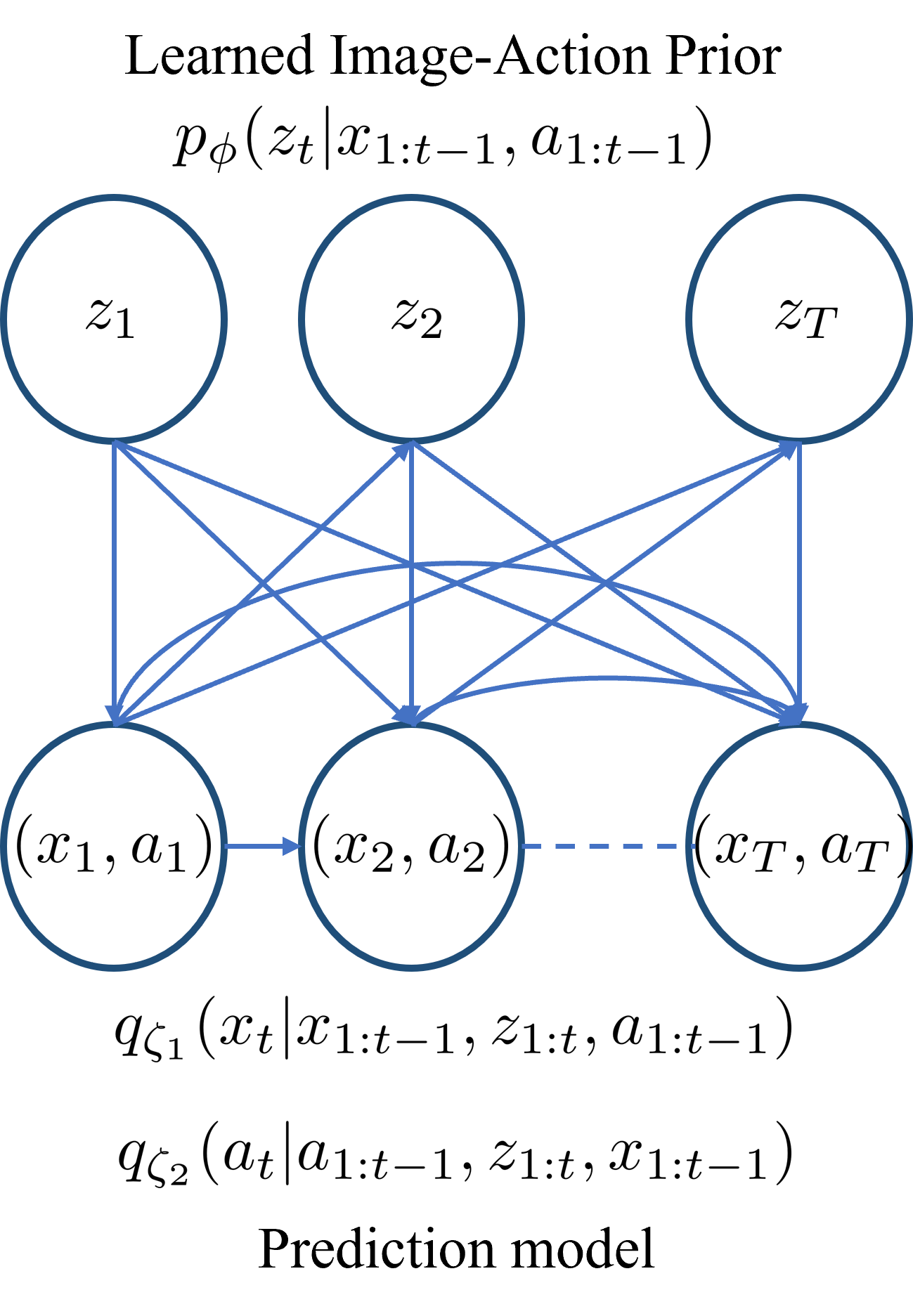}
        \caption{}
        \label{fig:vleap_state}
    \end{subfigure}
    \hfill
   \begin{subfigure}[t]{0.65\textwidth}
        \includegraphics[width=0.8\linewidth,height=6cm]{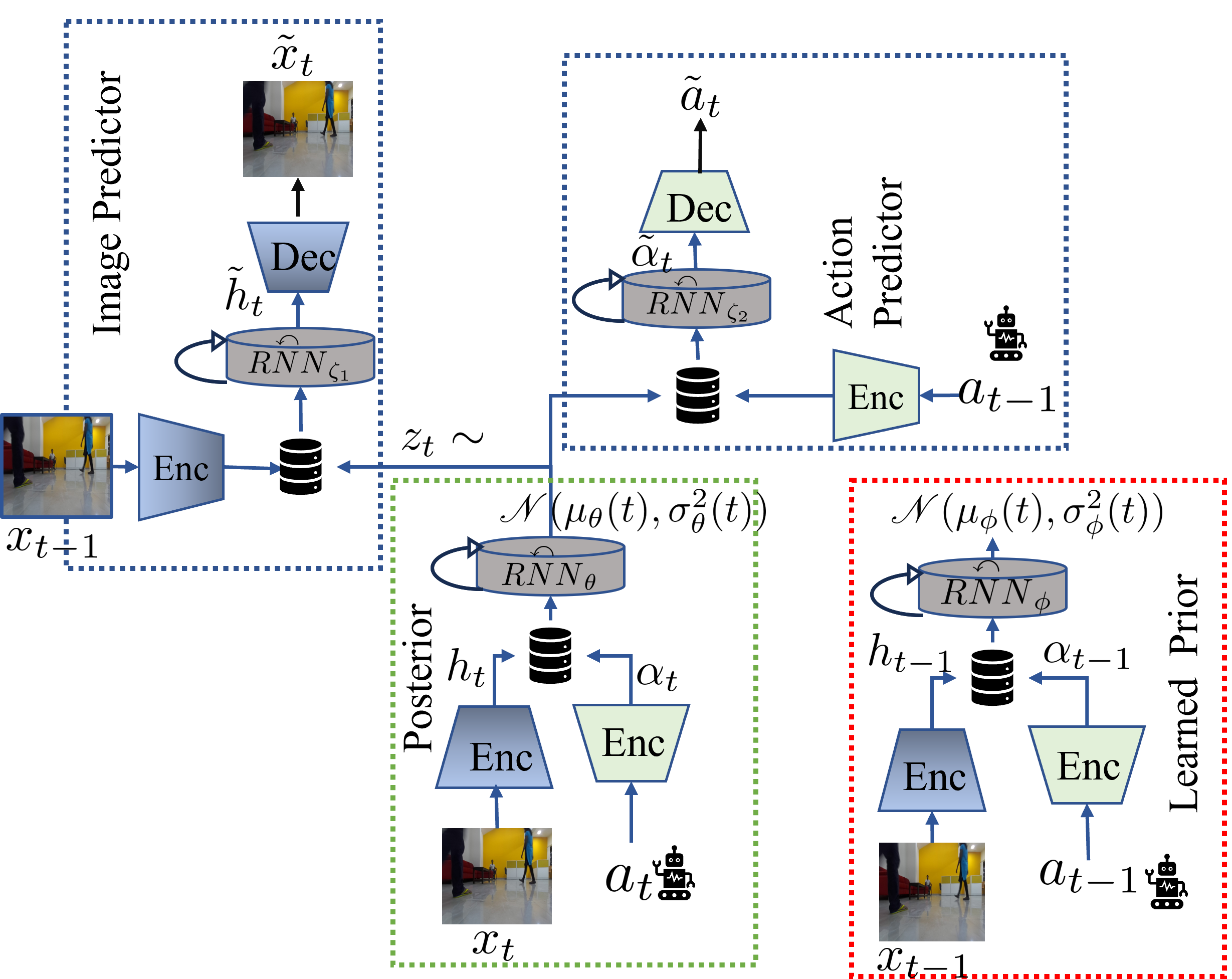}
        \caption{}
        \label{fig:vleap_block}
    \end{subfigure}
    \caption{ Fig \ref{fig:vleap_state} shows the state flow diagram and generation model for the VG-LeAP model with learned image-action prior $z_t$ that is dependent on the image action pair $(x_t,a_t)$. Fig \ref{fig:vleap_block} depicts the architecture of video generation with learned action prior model (\textcolor{red}{red} color dotted boxed portion) along with the posterior network in \textcolor{green}{green} color dotted boxed portion. At the time of inference only the prior model (\textcolor{red}{red} colored) is used. The prior and posterior latent models are trained using KL divergence loss.}
    \label{fig:enter-label}
\end{figure}
\subsection{Video generation with learnt action prior}\label{sec:vleap}
Our first model Video generation with learnt action prior or VG-LeAP is built on the principles of stochastic video generation in \cite{Denton}. However, unlike \cite{Denton} where only images were considered as the observed state of the stochastic process, we introduce the notion of image-action pair $(x_t,a_t)$ as an augmented state of the extended stochastic process that models the image frames as well as the action of the robot. In scenarios where the camera is moving, the observed image frames are influenced by the past actions or movements of the camera. Additionally, in many cases, the future actions of a robotic agent or a car (on which the camera is mounted) depend on the images observed, particularly when obstacle avoidance modules are integrated into the platform's motion planner. This interdependence between the image and action is also referred to as the partial observability problem in video prediction literature \cite{villegas,sarkar2021}. Thus modelling this process with the notion of system or robot action as a part of an extended state of the process provides a clear way of encapsulating these interdependent dynamics.  

We assume that the extended image-action pair $\chi_t=(x_t,a_t)$ is generated from a latent unknown process $p(z_t)$ of variable $z_t$ whose posterior is approximated with a recurrent neural architecture of parameter $\theta$ in the form $p_\theta(z_t|x_{1:t},a_{1:t})$. In order to learn this posterior distribution we employ a similar variational architecture as that of SVG-lp, however in our case we use the notion of extended state instead of just the image frames. We use the reparameterization trick from variational inference \cite{vae_kingma2014}, to approximate we approximate $p_\theta(z_t|x_{1:t},a_{1:t})$ as a Gaussion process such that  $z_t \sample \gN(\mu_\theta(z_t|\chi_{1:t-1}),\sigma_\theta(z_t|\chi_{1:t}))$ where $\mu$ and $\sigma$ denotes the mean and variance respectively. The state diagram of the learned image-action prior model in Fig \ref{fig:vleap_state} depicts this relationship between learned latent variable $z_t$ and observed image-action pair $(x_t,a_t)$ with connecting blue arrows. We also use a recurrent module parameterised by $\phi$ to learn the image-action prior $p_\phi(z_t|x_{1:t-1},a_{1:t-1})$ to use during inference when the current image $x_t$ and action $a_t$ is not available. The architecture of the network can be expressed as follows and is pictorially represented in Fig \ref{fig:vleap_block}:
\begin{equation}
    x_t\overset{Enc}{\longrightarrow} h_t, 
     \hspace{2em }  a_t\overset{Enc}{\longrightarrow} \alpha_t
    \label{eq:vleap_ht}
\end{equation}
\begin{equation}
    \mu_{\theta}(t),  \sigma_{\theta}(t)=\overset{\curvearrowleft}{RNN}_\theta (h_{0:t},\alpha_{0:t}), \hspace{2em }  
    z_t \sample \gN(\mu_\theta(t), \sigma^2_\theta(t))
    \label{eq:vleap_normal}
\end{equation}
\begin{equation}
    x_{t-1}\overset{Enc}{\longrightarrow} h_{t-1}, 
     \hspace{2em }  \tih_t=\overset{\curvearrowleft}{RNN}_{\zeta_{1}} (h_{0:t-1},z_{1:t})
    \label{eq:vleap_st}
\end{equation}
\begin{equation}
    a_{t-1}\overset{Enc}{\longrightarrow} \alpha_{t-1}, 
     \hspace{2em }  \talpha_t=\overset{\curvearrowleft}{RNN}_{\zeta_{2}} (\alpha_{0:t-1},z_{1:t})
    \label{eq:vleap_ut}
\end{equation}
\begin{equation}
    \tilde{x}_{t}\overset{Dec}{\longleftarrow} \tih_{t},\hspace{2em } \tilde{a}_{t}\overset{Dec}{\longleftarrow} \talpha_{t}
    \label{eq:vleap_xt}
\end{equation}
In \eqref{eq:vleap_ht} we encode image frames to a low dimensional manifold with $h_t$ and we map action data to a higher dimensional state of $\alpha_t$. These encoded features are then fed to the posterior estimation network for eventual sampling of $z_t$ in \eqref{eq:vleap_normal}. Please note the dependence in $z_t$ on the past data $(h_{0:t},\alpha_{0:t})$ arises from the recurrent LSTM components in the posterior network. This same dependence of the predicted image $\tih_t$ and action data $\talpha_t$ on the history of observed data $(h_{0:t-1},z_{0:t})$ and $(\alpha_{0:t-1},z_{0:t})$ in \eqref{eq:vleap_st} and \eqref{eq:vleap_ut} respectively are modelled with the LSTM components in the image and action predictor networks $\overset{\curvearrowleft}{RNN}_{\zeta_{1}}$ and $\overset{\curvearrowleft}{RNN}_{\zeta_{2}}$ respectively. Finally the generated image $\tilde{x}_{t}$ and action $\tilde{a}_{t}$ are decoded with their respective decoder architectures in \eqref{eq:vleap_xt}. The action conditioned prior $p_\phi(z_t|x_{1:t-1},a_{1:t-1})$ is learned as $\mu_{\phi}(t),  \sigma_{\phi}(t)=\overset{\curvearrowleft}{RNN}_\phi (h_{0:t-1},\alpha_{0:t-1})$.

\textbf{Loss:} We use a modified variational lower bound or ELBO loss in \eqref{eq:vleap_elbo} to train our VG-LeAP architecture.
\begin{equation}
\begin{aligned}
\underset{\theta,\phi, \zeta_1,\zeta_2}{\max}&\gL_{\theta,\phi, \zeta_1,\zeta_2}(x_{1,T},a_{1:T})=
\sum_{t=1}^{T}
   \big[\E_{p_\theta(z_{1:t}|x_{1:t},a_{1:t})}(\text{ln}q_{\zeta_1}(x_t|x_{1:t-1},z_{1:t})+\\ 
&\hspace{-0.5em}\beta_a\text{ln}q_{\zeta_2}(a_t|a_{1:t-1},z_{1:t}))
-\beta D_{KL} (p_\theta(z_t|x_{1:t},a_{1:t})||p_\phi(z_{t}|x_{1:t-1},a_{1:t-1}))\big]
\end{aligned}
\label{eq:vleap_elbo}
\end{equation}
The first and the third components in \eqref{eq:vleap_elbo} refer to the widely used reconstruction and KL divergence loss of variational frameworks \cite{Denton,villegasNeurIPS2019,chatterjee2021ICCV}, however, the second expectation term comes as a natural expansion of the extended state of $(x_t,a_t)$ that incorporates action $a_t$. In \eqref{eq:vleap_elbo}, $q_{\zeta_1}(x_t|\cdots)$ and $q_{\zeta_2}(a_t|\cdots)$ represents the likelihood functions of predicting $x_t$ and $a_t$ by $\overset{\curvearrowleft}{RNN}_{\zeta_{1}}$ and $\overset{\curvearrowleft}{RNN}_{\zeta_{2}}$ respectively and are approximated with the $L_{\text{p}}$ where $p \in \{1,2\}$ norm loses between the ground truth and predicted values. The hyper-parameters $\beta_a$ and $\beta$ are selected based on the numerical stability of the training and their selection process is discussed in detail in the supplementary material.
\begin{figure}[t]
    \centering
    \begin{subfigure}[t]{0.3\textwidth}
        \includegraphics[width=\linewidth]{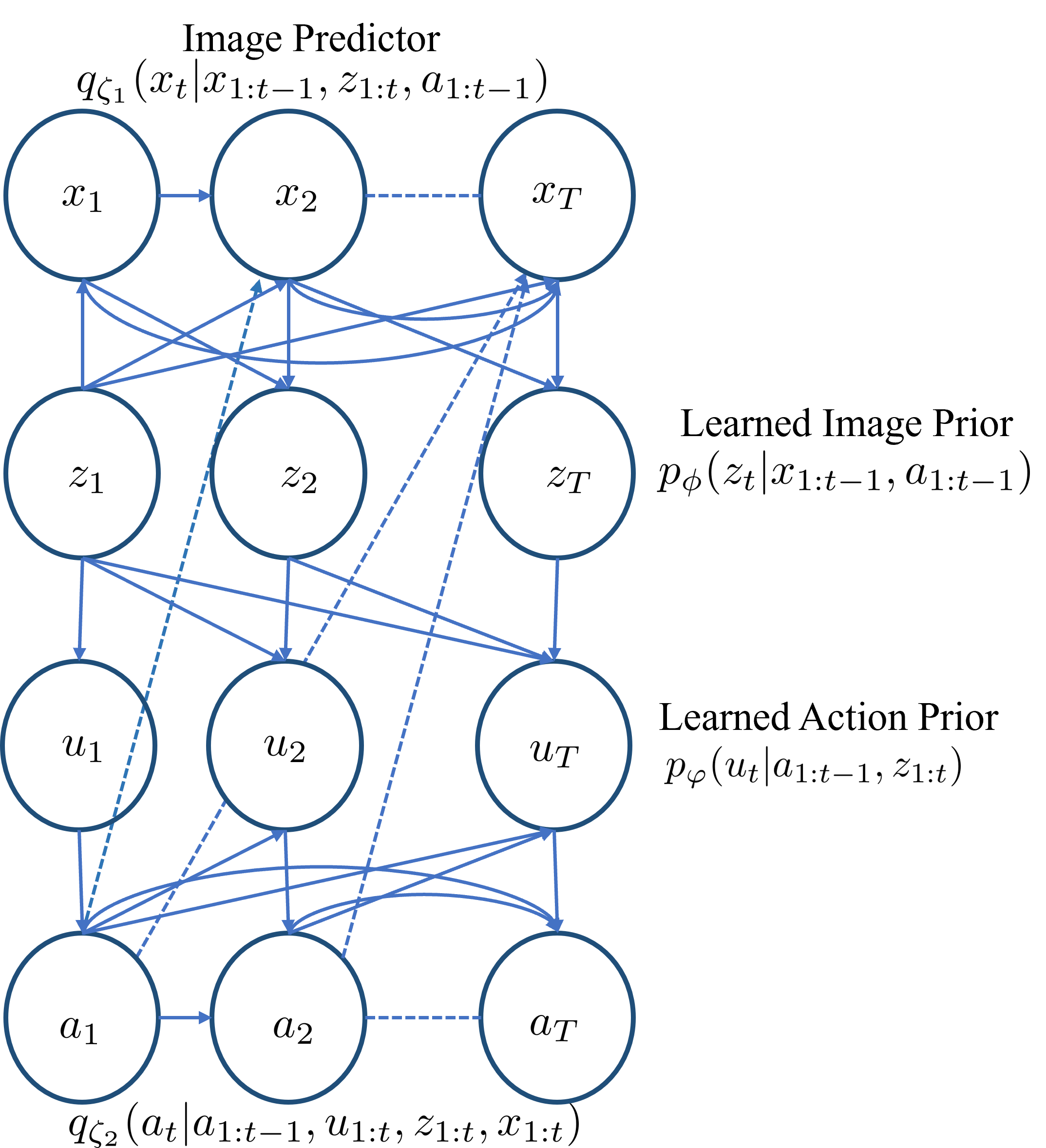}
        \caption{}
        \label{fig:causal_state}
    \end{subfigure}
    \hfill
   \begin{subfigure}[t]{0.65\textwidth}
        \includegraphics[width=\linewidth,height=7cm]{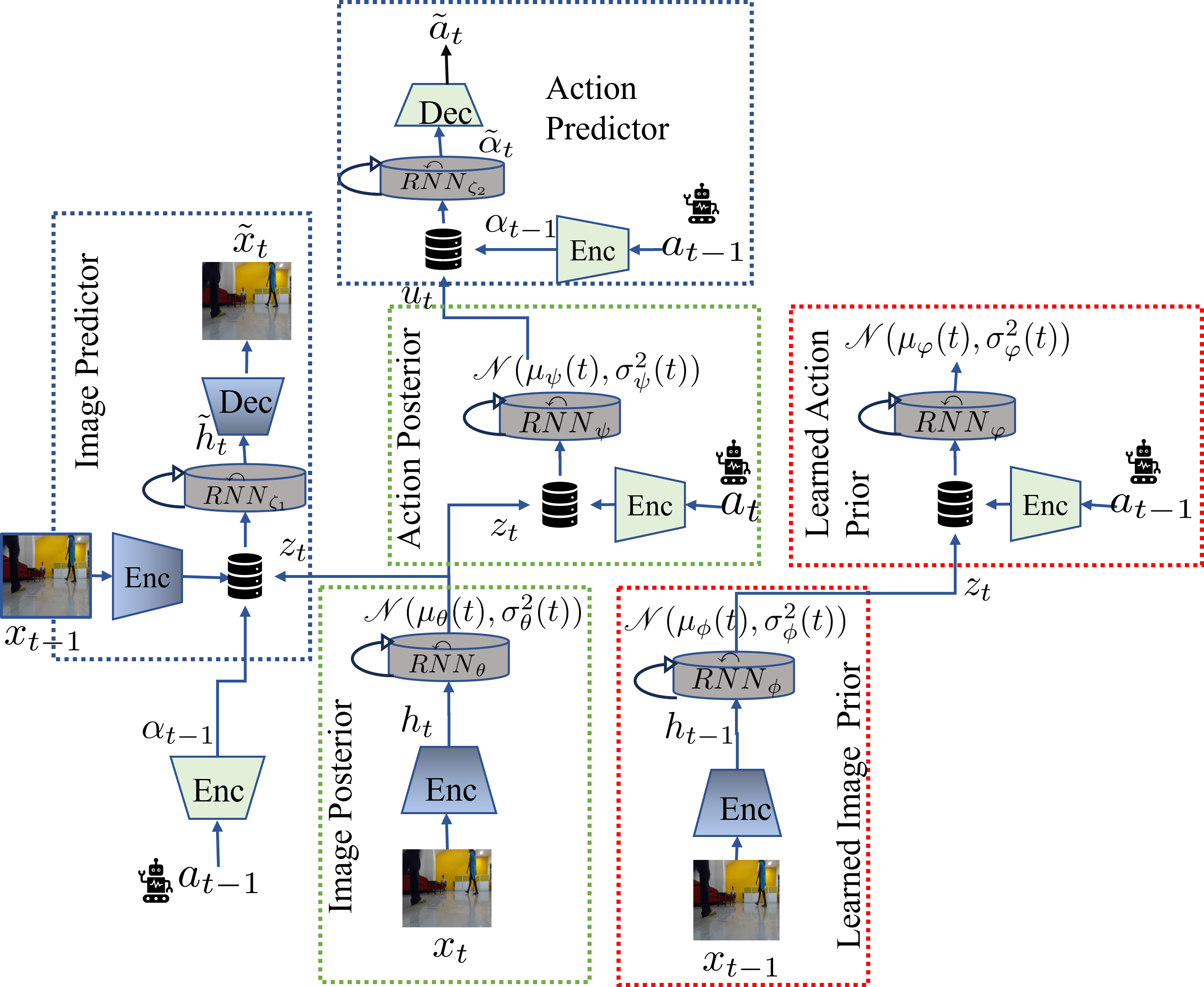}
        \caption{}
        \label{fig:causal_block}
    \end{subfigure}
    \caption{ Fig \ref{fig:causal_state} shows the state flow diagram and generation model for the Causal-LeAP model with learned action prior $u_t$ that is dependent on the learned image prior $z_t$ The forward causal relationship between image latent state $z_t$ image and action latent variable $u_t$ is depicted via the blue continuous connecting line. The dotted lines from $a_{t-1}$ to $x_t$ represent the dependency between past actions and future observed images. Fig \ref{fig:causal_block} depicts the architecture of video generation with learned both the action prior and image prior models (\textcolor{red}{red} colour dotted boxed portion). The posterior networks are shown in \textcolor{green}{green} colour-dotted boxed portions. At the time of inference only the prior models (\textcolor{red}{red} coloured) are used in the forward pass to sample $z_t$ and $u_t$ to generate $\tx_t$ and $\ta_t$. The prior and posterior latent models are trained using KL divergence loss.}
    \label{fig:cause-leap}
\end{figure}
\subsection{Causal video generation with learned action prior}\label{sec:causal-leap}
In our Causal-LeAP or Causal Learned Action Prior model, instead of treating the image-action pair $(x_t,a_t)$ as an extended state of the generative process, we adopt a causal approach. We assume a causal relationship between the action, $a_t$ taken by the moving platform or robot at time-step $t$ and the observed image frame  $x_t$. This approach aligns with most motion planning algorithms, which plan actions based on the current observed state, following Markovian models. Consequently, the action taken at time $t$ influences the image frame observed at $t+1$, $x_{t+1}$  and this causal chain continues sequentially. We assume that image $x_t$ is generated from a latent unknown process $p(z_t)$ of variable $z_t$ whose posterior is approximated with a recurrent neural architecture of parameter $\theta$ in the form $p_\theta(z_t|x_{1:t},a_{1:t-1})$ (This is similar to SVG-lp video generator part except for the conditioning of $z_t$ on past observed action $a_{1:t-1}$). Action $a_t$ is assumed to be generated from the latent process $p(u_t|x_t)$ where we have already observed $x_t$. We approximate the posterior of $u_t$ with an LSTM module of parameter $\psi$ of the form $p_\psi(u_t|a_{1:t},z_{1:t})$. The causal relationship between the image latent state $z_t$ and the action latent variable is depicted with the blue connecting lines in the state flow diagram in Fig \ref{fig:causal_state}. 

We reparameterize \cite{vae_kingma2014}, $p_\theta(z_t|x_{1:t},a_{1:t-1})$ and $p_\psi(u_t|a_{1:t},z_{1:t})$ as Gaussion processes such that  $z_t \sample \gN(\mu_\theta(z_t|x_{1:t},a_{1:t-1}),\sigma_\theta(x_{1:t},a_{1:t-1}))$ and $u_t \sample \gN(\mu_\psi(u_t|a_{1:t},z_{1:t}),\sigma_\psi(a_{1:t},z_{1:t}))$ respectively. Here $\mu$ and $\sigma$ signify the mean and variance of the distributions. With Causal-LeAP we train two recurrent modules parameterised by $\phi$ and $\varphi$ to learn the image prior $p_\phi(z_t|x_{1:t-1},a_{1:t-1})$ and causal action prior $p_\varphi(u_t|a_{1:t-1},z_{1:t-1})$ respectively. $p_\phi(z_t|\cdots)$ and $p_\varphi(u_t|\cdots)$ are used at the time of inference when the current image $x_t$ and action $a_t$ are not available. The architecture of the network can be expressed as follows and is depicted in Fig \ref{fig:causal_block}:
\begin{equation}
    x_t\overset{Enc}{\longrightarrow} h_t, 
     \hspace{2em }  a_t\overset{Enc}{\longrightarrow} \alpha_t
    \label{eq:causal_ht}
\end{equation}
\begin{equation}
    \mu_{\theta}(t),  \sigma_{\theta}(t)=\overset{\curvearrowleft}{RNN}_\theta (h_{1:t}), \hspace{2em }  
    z_t \sample \gN(\mu_\theta(t), \sigma^2_\theta(t))
    \label{eq:causal_normal}
\end{equation}
\begin{equation}
    \mu_{\psi}(t),  \sigma_{\psi}(t)=\overset{\curvearrowleft}{RNN}_\psi (\alpha_{1:t},z_{1:t}), \hspace{2em }  
    u_t \sample \gN(\mu_\psi(t), \sigma^2_\psi(t))
    \label{eq:causal_ut}
\end{equation}
\begin{equation}
    x_{t-1}\overset{Enc}{\longrightarrow} h_{t-1}, 
     \hspace{2em }  \tih_t=\overset{\curvearrowleft}{RNN}_{\zeta_{1}} (h_{1:t-1},z_{1:t},\alpha_{1:t-1})
    \label{eq:causal_st}
\end{equation}
\begin{equation}
    a_{t-1}\overset{Enc}{\longrightarrow} \alpha_{t-1}, 
     \hspace{2em }  \talpha_t=\overset{\curvearrowleft}{RNN}_{\zeta_{2}} (\alpha_{1:t-1},u_{1:t})
    \label{eq:causal_alphat}
\end{equation}
\begin{equation}
    \tilde{x}_{t}\overset{Dec}{\longleftarrow} \tih_{t},\hspace{2em } \tilde{a}_{t}\overset{Dec}{\longleftarrow} \talpha_{t}
    \label{eq:causal_xt}
\end{equation}
We encode image frames and actions to  $h_t$ and $\alpha_t$ respectively in \eqref{eq:causal_ht} which is similar to \eqref{eq:vleap_ht} in \ref{sec:vleap}. The encoded image vectors $h_t$ are then fed to the posterior estimation network $\overset{\curvearrowleft}{RNN}_\theta$ for the eventual sampling of $z_t$ in \eqref{eq:causal_normal}. Please note the dependence in $z_t$ on the past data $(h_{1:t},\alpha_{1:t-1})$ in \eqref{eq:causal_normal} arises from the recurrent LSTM components in the posterior network and in \eqref{eq:causal_normal} $z_t$ does not depend upon $a_t$ like in \eqref{eq:vleap_normal}. Equation \ref{eq:causal_ut} encapsulates the causal relationship between $x_t$ and $a_t$ as the image latent variable is fed to $\overset{\curvearrowleft}{RNN}_\psi$ to generate $u_t$. The predicted image $\tih_t$ and action data $\talpha_t$ are dependent on the history of observed data $(h_{1:t-1},z_{1:t},\alpha_{1:t-1})$ and $(\alpha_{1:t-1},u_{1:t})$ in \eqref{eq:causal_st} and \eqref{eq:causal_alphat} respectively and these dependencies are modelled with LSTM modules in the image and action predictor networks $\overset{\curvearrowleft}{RNN}_{\zeta_{1}}$ and $\overset{\curvearrowleft}{RNN}_{\zeta_{2}}$ respectively. Finally the generated image $\tilde{x}_{t}$ and action $\tilde{a}_{t}$ are decoded with their respective decoder architectures in \eqref{eq:causal_xt}. The action conditioned image prior $p_\phi(z_t|x_{1:t-1},a_{1:t-1})$ is learned as $\mu_{\phi}(t),  \sigma_{\phi}(t)=\overset{\curvearrowleft}{RNN}_\phi (h_{1:t-1})$ and the causal learned action prior $p_\varphi(u_t|a_{1:t-1},z_{1:t})$ is learned as $\mu_{\varphi}(t),  \sigma_{\varphi}(t)=\overset{\curvearrowleft}{RNN}_\varphi (\alpha_{1:t-1},z_{1:t-1})$.

\textbf{Loss:} We derive our variational lower bound or ELBO loss in \eqref{eq:causal_elbo} to train our Causal-LeAP architecture.
\begin{equation}
\begin{aligned}
\underset{\theta,\phi,\psi,\varphi, \zeta_1,\zeta_2}{\max}&\gL_{\theta,\phi,\psi,\varphi, \zeta_1,\zeta_2}(x_{1,T},a_{1:T})=
\sum_{t=1}^{T}\big[\E_{p_\theta(z_{1:t}|x_{1:t})}\text{ln}q_{\zeta_1}(x_t|x_{1:t-1},z_{1:t},a_{1:t-1})-\\ 
&\hspace{-3.0em}\beta D_{KL} (p_\theta(z_t|x_{1:t})||p_\phi(z_{t}|x_{1:t-1}))+\beta_a\E_{p_\psi(u_{1:t}|z_{1:t},a_{1:t})}\text{ln}q_{\zeta_2}(a_t|a_{1:t-1},u_{1:t},z_{1:t})\\
&\hspace{12em}-\gamma D_{KL} (p_\psi(u_t|a_{1:t},z_{1:t})||p_\varphi(u_{t}|a_{1:t-1},z_{1:t}))\big]
\end{aligned}
\label{eq:causal_elbo}
\end{equation}
In \eqref{eq:causal_elbo}, the first two components represent the conventional reconstruction and KL divergence loss components of variational learning. The third and fourth components come from maximizing the log-likelihood of $p(a_t|x_t)$ or $\text{ln}p(a_t|x_t)$.
The likelihood functions $q_{\zeta_1}(x_t|\cdots)$ and $q_{\zeta_2}(a_t|\cdots)$ for predicting $x_t$ and $a_t$ by $\overset{\curvearrowleft}{RNN}_{\zeta_{1}}$ and $\overset{\curvearrowleft}{RNN}_{\zeta_{2}}$ respectively are approximated with the $L_{\text{p}}$ where $p \in \{1,2\}$ norm loses between the ground truth and predicted values.
The hyper-parameter $\gamma$ relating to the KLD loss associated with the action prior function is chosen according to the numerical stability of the problem. In this case, the action predictor is a much smaller model compared to the image predictor and thus tends to converge much quicker which can lead to numerical instability in case of large learning rates or very small $\beta$ values. The selection criteria for all the three hyper-parameters $\beta,\beta_a$ and $\gamma$ are discussed in the supplementary.
\subsection{Random Action-Frame Conditioned Flow Integrating video generation model}
Our diffusion model Random Action-Frame Conditioned Flow Integrating video generator or RAFI is based on the sparsely conditioned flow matching model of RIVER by \cite{river2023ICCV}. Like RIVER, we also encode our image states in the latent space of a pre-trained VQGAN \cite{vqgan2021}. However, unlike RIVER, we then join the latent image state $z_t$ from the VQGAN network with the action vectors to generate the extended image-action state $\tz_t$ as shown in the fourth step in Algo. \ref{alg:vfm-river}. Following $\tz_t$, we follow the steps similar to RIVER to train the flow vector regressor \cite{flowMatching2023ICLR} using gradient descent. The step by step algorithm for RAFI is given in Algo. \ref{alg:vfm-river} and for more details about the implementation of the algorithm please refer to our Models and Training section of supplementary. 
\begin{algorithm}[t]
\caption{Training Procedure for RAFI}
\label{alg:vfm-river}
\begin{algorithmic}[1]
\REQUIRE Dataset of image, action pair sequence  $\gD$, number of training iteration $N$
\FOR{$i$ in range$(1, N)$}
    \STATE Sample a sequence of image frames $x_{1:T}$ and corresponding action sequence $a_{1:T}$ from the dataset $\gD$
    \STATE Encode all the images frames $x_{1:T}$ with a pre-trained VQGAN to obtain $z_{1:T}$
    \STATE For each $x_t$, concat action $a_t$ as additional channels to the output of VQGAN to get $\tz_t$ 
    \STATE Choose a random target frame $\Tilde{z}_\tau$, $\tau \in \{3, \ldots, T\}$
    \STATE Sample a timestamp $t \sim U[0, 1]$
    \STATE Sample a noisy observation $\nu \sim p_t(\Tilde{z} \mid \Tilde{z}^\tau)$
    \STATE Calculate target vector filed $\gU_t(\nu \mid \Tilde{z}^\tau)$
    \STATE Sample a condition frame $\Tilde{z}^c$, $c \in \{1, \ldots, \tau - 2\}$
    \STATE Update the parameters $\theta$ of the flow vector field regressor $v_t$ with gradient descent:
    \begin{equation}
        \nabla_\theta \|v_t(\nu \mid \Tilde{z}^{\tau-1}, \Tilde{z}^c, \tau - c; \theta) - \gU_t(\nu \mid \Tilde{z}^\tau)\|^2
        \label{eq:vfm-river-loss}
    \end{equation}
\ENDFOR
\end{algorithmic}
\end{algorithm}
\section{Dataset and Experiments:}
\subsection{RoAM dataset:}
RoAM or Robot Autonomous Motion dataset is a synchronised and timestamped image-action pair sequence dataset, recorded with a Turtlebot3 Burger robot with a Zed mini stereo camera. The dataset was first introduced by \cite{acpnet2023} to establish the connection between the generated image frames and the robot action data. RoAM is recorded indoors capturing corridors, lobby spaces, staircases, and laboratories featuring frequent human movement like walking, sitting down, getting up, standing up, etc. However, the original RoAM in \cite{acpnet2023} only contained 25 long video sequences out of which only 20 were used for training. This dataset was very small for training stochastic frameworks such as SVG or RIVER. So we have recorded another 25 long video frames in various indoor environments with varying lighting conditions. Now the dataset is segregated into 45 long training video sequences and 5 sequences are kept for testing. Using these 45 video sequences we have used the Tensorflow \cite{tensorflow2015} Dataset API to generate 3,07,200 video sequences of length 40 and image size $64\times64\times4$. It also contains the corresponding action values from the robot's motion to capture the movement of the camera. The dimension of the action data in RoAM is $m=2$  featuring forward velocity along the body $x$-axis and turn rate about the body $z$-axis of the robot's centre of mass.

More details on the process of RoAM dataset recording and our training pipeline are discussed in the Dataset section of our supplementary file.
\subsection{Experimental Setup:}
Out of the 25 frames in each sequence, we used 5 random frames to condition our networks VG-LeAP, Causal-LeAP and SVG on the past data. In the case of RIVER and RAFI we only condition them on the past two frames and in all 5 models we generated the next 10 frames in the future during training. In order to test the networks, we have created 1024 randomly generated video sequences of length 40 from the origin 5 test sequences in RoAM and tested all the 5 networks against the quantitative performance metrics such as: Peak Signal-to-Noise Ratio (PSNR), VGG16 Cosine Similarity \cite{VGG16}, and Fréchet Video Distance (FVD) \cite{FVD} and Learned Perceptual Image Patch Similarity or LPIPS  metric\cite{lpips}. Among these metrics, FVD is based on the Fréchet Inception Distance (FID) that is commonly used for evaluating the quality of images from generative frameworks and measures the spatio-temporal distribution of the generated videos as a whole, with respect to the ground truth. We use VGG16 cosine similarity index, LPIPS and PSNR for frame-wise evaluation. The VGG16 cosine similarity index uses the pre-trained VGG16 network \cite{VGG16} to measure the cosine similarity between the generated and ground truth video frames. Recently perceptual similarity metric LPIPS \cite{lpips} which uses pretrained AlexNet as its image feature generator, has emerged as a popular measure \cite{slrvp2020} for its human-like perception of similarity between two image frames. In case of VGG16 Cosine Similarity and PSNR values, closer resemblance to the ground truth images is indicated by higher values whereas in LPIPS and FVD scores, superior performance is associated with lower values.
\section{Results and Discussion:} \label{sec:results}
During inference, we tested all our 6 models: VG-LeAP, Causal LeAP, RAFI along with SVG-lp, RIVER and ACPNet on predicting 20 future frames conditioned on the past 5 image frames and the plots for LPIPS, VGG Cosine Similarity and PSNR are shown in Fig \ref{fig:final_quan}. From all the figures in Fig \ref{fig:final_quan}, we can easily see that our Causal-LeAP and VG-LeAP models easily outperform the SVG-lp model on RoAM dataset. While all these three models Causal-LeAP, VG-LeAP and SVG-lp share similar image predictor architectures, it can be easily concluded that the improved behaviour is a direct result of modelling the combined image-action dynamics in the case of VG-LeAP and Causal-LeAP. Comparing the behaviour between SVG and VG-LeAP, where both the networks share almost identical architecture and size of the parametric space, we can very clearly see that VG-LeAP outperforms SVG in Fig. \ref{fig:final_lpips}, Fig \ref{fig:final_vgg}, and Fig \ref{fig:final_psnr}. The mean FVD score of VG-LeAP is around 481.15 which is better than the 539.29 from SVG.

In between VG-LeAP and Causal-LeAP, Causal-LeAP outperforms VG-LeAP in almost every quantitative metric (Fig. \ref{fig:final_quan}) except for FVD score shown in Table \ref{tab:fvd_exd} (Causal LeAP has a FVD score of 514.65). In the case of our two diffusion-based models RIVER and RAFI, we can see that both of them initially perform much better than Causal-LeAP and VG-LeAP (Fig \ref{fig:final_lpips},\ref{fig:final_vgg}), however as time increases their performance gets worse. However, in terms of FVD scores, RIVER and RAFI generate the best results with mean scores of 284.46 and 288.23 respectively (Table \ref{tab:fvd_exd}).

\begin{figure}[t]
    \centering
    \begin{subfigure}[t]{0.3\textwidth}
        \includegraphics[width=\linewidth]{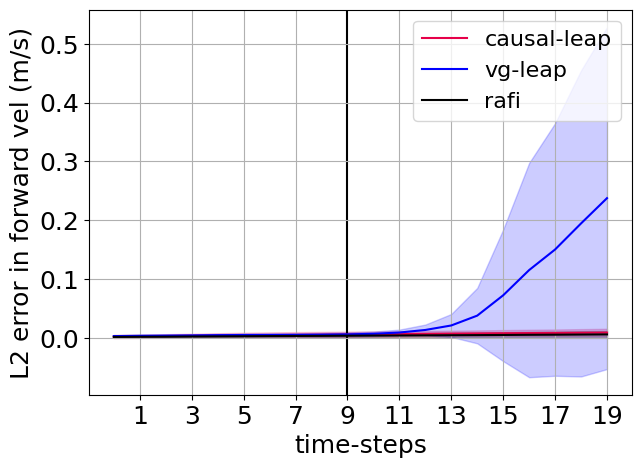}
        \caption{}
        \label{fig:final_at1_f}
    \end{subfigure}
    \hfill
   \begin{subfigure}[t]{0.32\textwidth}
        \includegraphics[width=\linewidth]{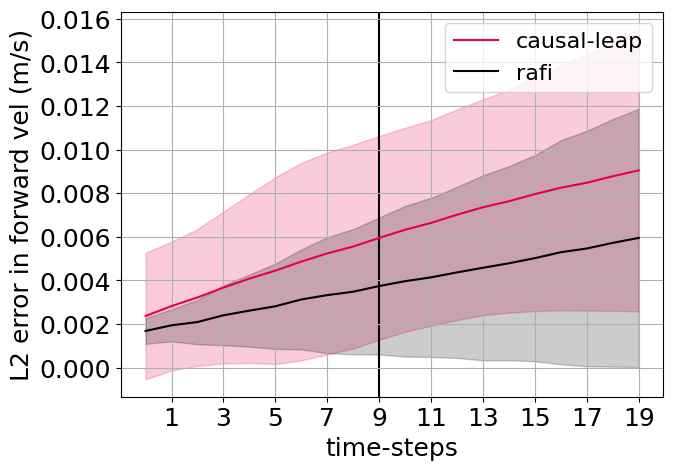}
        \caption{}
        \label{fig:final_at1_r}
    \end{subfigure}
    \hfill
    \begin{subfigure}[t]{0.32\textwidth}
        \includegraphics[width=\linewidth]{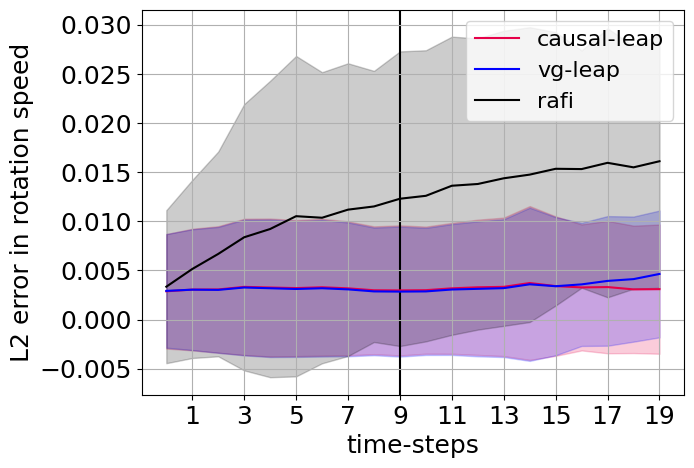}
        \caption{}
        \label{fig:final_at2}
    \end{subfigure}
    \caption{ Fig. \ref{fig:final_at1_f},\ref{fig:final_at1_r}  and \ref{fig:final_at2} show the quantitative L$_2$ norm error between the predicted action values and the ground truth for Causal-LeAP, VG-LeAP and RAFI. In Fig. \ref{fig:final_at1_f} we have shown the error in the normalised forward velocity between Causal-LeAP, VG-LeAP and RAFI. In this case, it can be seen that even though initially all the 3 models perform similarly to each other, as time increases, Vg-LeAP produces much more noisy and erroneous predictions compared to the other two models. Fig. \ref{fig:final_at1_r} shows that when compared between Causal-LeAP and RAFI, RAFI performs much better in time when it comes to predicting the forward velocity. Fig. \ref{fig:final_at2} shows that in the case of angular rotation or turn rate, Causal-LeAP provides the best predictions and RAFI performs the worst. }
    \label{fig:final_at}
\end{figure}

We have also plotted the comparative L$_2$ norm errors in the predicted action data from Causal-LeAP, VG-LeAP and RAFI in Fig \ref{fig:final_at} and here we see that RAFI performs better than all the other two models when it comes to predicting the forward velocity. However, for turn rate, RAFI does not provide reliable predictions as compared to both Causal-LeAP and VG-LeAP in Fig \ref{fig:final_at2}. It produces erroneous turn rates which we believe also has an adverse effect in the generated images by RAFI. Since RAFI treats Image-Action as an extended state, the rotation in action also results in generated images that are rotated and thus the prediction accuracy decreases. 

We can also observe from Fig \ref{fig:final_at} that until $t=13$, both VG-LeAP and Causal-LeAP generated similar $\text{L}_2$ norm error in forward velocity, however after $t=12$, in the case of VG-LeAP, the action error starts to accumulate exponentially whereas, in case of Causal-LeAP, the action error loss remains almost constant. 

The additional KL divergence term from the action network in the loss function for Causal-LeAP also acts as an additional regularize to the causal training process which may lead to numerical instability during training. However, we found keeping a small $\gamma$ value in \eqref{eq:causal_elbo} can easily reduce the scenarios. The numerical instability in the training of the Causal-LeAP model also arises from the disproportionate size of the parametric spaces of the Image and the Action prediction networks. Since the Action prediction framework converges faster than the Image prediction module, we found keeping a low $\gamma$ is essential for the training.

 
In the case of ACPNet which is the only deterministic model in our empirical study, we have found that even though ACPNet initially starts generating good prediction (\ref{fig:final_lpips},\ref{fig:final_vgg}), it quickly suffers from blurring effects that are common in deterministic architectures. The blurring effect is also the reason why ACPNet produces the best PSNR result \cite{lpips,slrvp2020} among all the 6 models. The FVD score for ACPNet is 908  (Table \ref{tab:fvd_exd}).

Additional results on generated raw image frames from all the 6 networks can be found on our project page: \url{https://meenakshisarkar.github.io/Motion-Prediction-and-Planning/dataset/}

\begin{figure}[t]
    \centering
    \begin{subfigure}[t]{0.32\textwidth}
        \includegraphics[width=\linewidth]{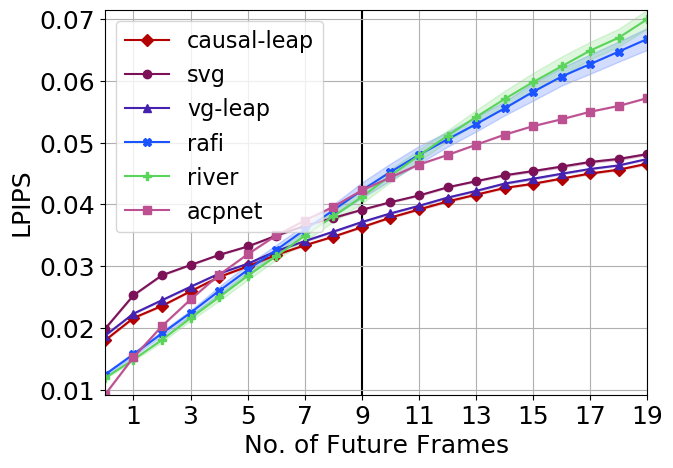}
        \caption{}
        \label{fig:final_lpips}
    \end{subfigure}
    \hfill
   \begin{subfigure}[t]{0.32\textwidth}
        \includegraphics[width=\linewidth]{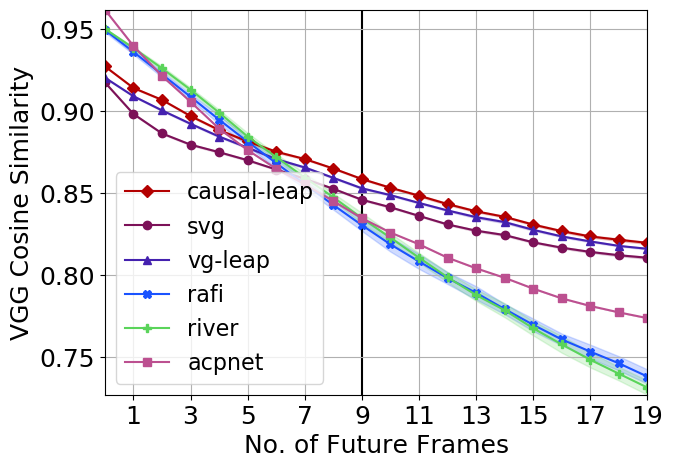}
        \caption{}
        \label{fig:final_vgg}
    \end{subfigure}
    \hfill
    \begin{subfigure}[t]{0.31\textwidth}
        \includegraphics[width=\linewidth]{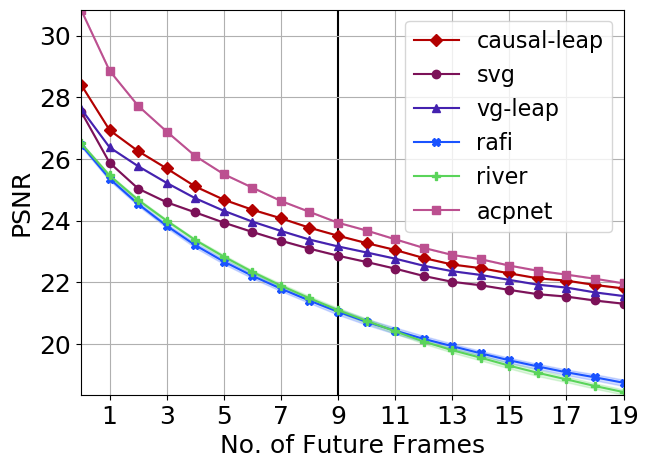}
        \caption{}
        \label{fig:final_psnr}
    \end{subfigure}
    \caption{ Fig. \ref{fig:final_lpips} (lower is better),\ref{fig:final_vgg}(higher is better)  and \ref{fig:final_psnr}(higher is better) showing the quantitative performance of Causal-LeAP, VG-LeAP, SVG (SVG-lp), RAFI,SRVP, and ACPNet for 20 different sampling on predicting 20 future image frames from past 5 conditioning frames. In all the quantitative performance metrics, Causal-LeAP model outperforms the other 5. In the case of LPIPS values for RAFI and ACPNet, we can see that both these models start much better than Causal-Leap, however as time passes, both start performing much worse than LeAP models. However, the reason for this performance degradation is completely different in the case of these two models as explained in Sec. \ref{sec:results} }
    \label{fig:final_quan}
\end{figure}
\begin{table}[!htb]
    \caption{FVD Score}
    \begin{subtable}{\linewidth}
      \centering
        \begin{tabular}{ll}
            \hline
 Model & Score  \\ \hline
 Causal-LeAP & 514.65 $\pm$ 3.37 \\ 
 VG-LeAP & 481.15 $\pm$ 2.39 \\
 SVG-lp & 539.29 $\pm$ 1.94 \\ 
 \textbf{RIVER (BEST)} & \textbf{284.46 $\pm$ 3.21} \\
 \textbf{RAFI } & \textbf{288.23 $\pm$ 4.39} \\
  ACPNET & 908.36 \\
 \hline
\end{tabular}\label{tab:fvd1}
    \end{subtable}
\label{tab:fvd_exd}
\end{table}
\section{Concusion:}
We have presented three new stochastic video generative frameworks based on the mathematical premise of incorporating action into the video generation process. We have also established a causal relationship between the image and camera actions in the partially observable scenarios where the camera is moving with our Causal-LeAP model and have shown with our detailed empirical studies that not only image-action models improve the efficacy of the prediction framework but also provides a way to learn and model the system dynamics by simply observing and modelling the interaction between the image-action pair. Our Causal model learned an action prior conditioned on the latent image state $p_\varphi(u_t|a_{1:t-1},z_{1:t})$ which has direct applications to the field of robotics and RL. Our model diffusion model RAFI also shows how easily we can extend the concepts of image-action state pair to existing diffusion approaches leading to good results.
{
    \small
    \bibliographystyle{bib}
    \bibliography{pami}
}
\appendix

\section{Appendix}
\subsection{Variational Lower Bound for Video Generation with Learned Action Prior}
Here we are trying to maximize the joint likelihood of $(x_{1:t},a_{1:T})$ which is equivalent to maximizing $\text{ln}q_{\zeta}(x_{1:T},a_{1:T})$ or $\text{ln}q_{\zeta}(x,a)$ for better readability. Let's assume $z=[z_1,\cdots,z_T]$  denotes the latent $z$ variable across all the time-steps and they are independent of each other across time.  
\begin{align}
\text{ln}q_{\zeta}(x_{1:T},a_{1:T})&\equiv \text{ln}q_{\zeta}(x,a)=\ln \int_{z} q_{\zeta} (x,a|z)p(z)\\
&=\ln \int_{z}q_{\zeta}(x,a|z)p(z)\frac{p_\theta(z|x,a)}{p_\theta(z|x,a)}\\
&=\ln \left(\E_{p_\theta(z|x,a)} q_{\zeta}(x,a|z)\frac{p(z)}{p_\theta(z|x,a)}\right)\\
&\geq \E_{p_\theta(z|x,a)} \left(\ln q_{\zeta}(x,a|z)\frac{p(z)}{p_\theta(z|x,a)}\right)\\
&=\E_{p_\theta(z|x,a)}\ln q_{\zeta}(x,a|z)-\E_{p_\theta(z|x,a)} \left(\ln\frac{p_\theta(z|x,a)}{p(z)}\right)\\
&=\E_{p_\theta(z|x,a)}\ln q_{\zeta}(x,a|z)- D_{KL}(p_\theta(z|x,a)||p(z))
\label{eq:pf_vleap_elbo6}
\end{align}
Given that we assumed in our model that $x$ and $a$ and conditionally independent given $z$, thus $q_{\zeta}(x,a|z)=q_{\zeta_1}(x|z) q_{\zeta_2}(a|z)$, where $\zeta=\{\zeta_{1},\zeta_{2}\}$. Thus \eqref{eq:pf_vleap_elbo6} can be written as :
\begin{align}
\text{ln}q_{\zeta}(x,a)&\geq\E_{p_\theta(z|x,a)}\ln q_{\zeta_1}(x|z)+\E_{p_\theta(z|x,a)}\ln q_{\zeta_2}(a|z)- D_{KL}(p_\theta(z|x,a)||p(z))
\label{eq:pf_vleap_elbo}
\end{align}
Similar to SVG, we use RNN architectures in VG-LeAP to recursively predict image frames and at each time-step $t$, $\overset{\curvearrowleft}{RNN}_{\zeta_{1}}$ takes the encoded past image $x_{t-1}$ and $z_t$ as input. With the recursive behaviour of $\overset{\curvearrowleft}{RNN}_{\zeta_{1}}$, we can express $\ln q_{\zeta_1}(x|z)$ as:
\begin{align}
   \ln q_{\zeta_1}(x_{1:T}|z_{1:T})&\equiv \ln q_{\zeta_1}(x|z)=\ln\prod_{t}q_{\zeta_1}(x_{t}|x_{1:t-1},z_{1:T})\\
   &=\sum_t \ln q_{\zeta_1}(x_{t}|x_{1:t-1},z_{1:t})
   \label{eq:pf_vleap_elbox}
\end{align}
In the case of action predictor in VG-LeAP, we use a similar RNN architecture $\overset{\curvearrowleft}{RNN}_{\zeta_{2}}$ which takes the past action $a_{t-1}$ and $z_t$ as input. Thus $ln q_{\zeta_2}(a|z)$ can be expressed as: 
\begin{align}
   \ln q_{\zeta_2}(a_{1:T}|z_{1:T})&\equiv \ln q_{\zeta_2}(a|z)=\ln\prod_{t}q_{\zeta_2}(a_{t}|a_{1:t-1},z_{1:T})\\
   &=\sum_t \ln q_{\zeta_2}(a_{t}|a_{1:t-1},z_{1:t})
   \label{eq:pf_vleap_elboa}
\end{align}
In the case of the posterior and learned prior networks $\overset{\curvearrowleft}{RNN}_{\theta}$ and $\overset{\curvearrowleft}{RNN}_{\phi}$ respectively, we recursively feed the action $a_t$ and image $x_t$ to approximate $z_t$ (in case of learned prior we feed $x_{t-1}$ and $a_{t-1}$). Cause $z_t$s are independent across time,  $p_\theta(z|x,a)$ can be expressed as:
\begin{align}
   p_\theta(z_{1:T}|x_{1:T},a_{1:T})&\equiv  p_\theta(z|x,a)=\prod_{t}p_\theta(z_t|x_t,a_t)
\end{align}
We assume the extended image-action state as $\chi=(x,a)$ for better readability and compact expressions in long equations. Since $z_t$s are independent across time, we can rewrite $D_{KL}(p_\theta(z|x,a)||p(z))$ or $D_{KL}(p_\theta(z|\chi)||p(z))$as:
\begin{align}
   D_{KL}(p_\theta(z|x,a)||p(z))&\equiv D_{KL}(p_\theta(z_{1:T}|x_{1:T},a_{1:T})||p(z_{1:T}))=\int_{z}p_\theta(z|x,a)\ln \frac{p_\theta(z|\chi)} {p(z)}\\
   &=\int_{z_1}\cdots \int_{z_T}p_\theta(z_1|\chi_1)\cdots  p_\theta(z_T|\chi_{1:T})\ln \frac{p_\theta(z_1|\chi_1)\cdots  p_\theta(z_T|\chi_{1:T})}{p(z_1)\cdots p(z_T)}\\
   &=\int_{z_1}\cdots \int_{z_T}p_\theta(z_1|\chi_1)\cdots  p_\theta(z_T|\chi_{1:T})\sum_t \ln \frac{p_\theta(z_t|\chi_{1:t})}{p(z_t)}\\
   &=\sum_t \int_{z_1}\cdots \int_{z_T}p_\theta(z_1|\chi_1)\cdots  p_\theta(z_T|\chi_{1:T}) \ln \frac{p_\theta(z_t|\chi_{1:t})}{p(z_t)}
\label{eq:pf_vleap_kld1}
\end{align}
Since $\int_z p_\theta(z)=1$ we can further simplify \eqref{eq:pf_vleap_kld1} as
\begin{align}
   D_{KL}(p_\theta(z|x,a)||p(z))&=\sum_t \int_{z_t}p_\theta(z_t|\chi_{1:t}) \ln \frac{p_\theta(z_t|\chi_{1:t})}{p(z_t)}\\
   &=\sum_t  D_{KL} (p_\theta(z_t|\chi_{1:t})||p(z_t))
   =\sum_t  D_{KL} (p_\theta(z_t|x_{1:t},a_{1:t})||p(z_t))
\label{eq:pf_vleap_kld}
\end{align}

Thus combining \eqref{eq:pf_vleap_elbox}, \eqref{eq:pf_vleap_elboa}and \eqref{eq:pf_vleap_kld}, with \eqref{eq:pf_vleap_elbo} we get the variational lower bound as:
\begin{align}
\text{ln}q_{\zeta}(x,a)&\geq\E_{p_\theta(z|x,a)}\ln q_{\zeta_1}(x|z)+\E_{p_\theta(z|x,a)}\ln q_{\zeta_2}(a|z)- D_{KL}(p_\theta(z|x,a)||p(z))\\
&=\sum_t [\E_{p_\theta(z_{1:t}|x_{1:t},a_{1:t})}(\ln q_{\zeta_1}(x_{t}|x_{1:t-1},z_{1:t})+\ln q_{\zeta_2}(a_{t}|a_{1:t-1},z_{1:t}))\nonumber\\
&\hspace{16em}-D_{KL} (p_\theta(z_t|x_{1:t},a_{1:t})||p(z_t))]
\label{eq:pf_vleap_elbo_final}
\end{align}
\subsection{Variational Lower Bound for Causal Video Generation with Learned Action Prior}
Here we are trying to maximize the joint likelihood of $(x_{1:t},a_{1:T})$ which is equivalent to maximizing $\text{ln}q_{\zeta}(x_{1:T},a_{1:T})$ or $\text{ln}q_{\zeta}(x,a)$ for better readability. Let's assume $z=[z_1,\cdots,z_T]$  denotes the image latent $z$ variable across all the time-steps and $u=[u_1,\cdots,u_T]$  denotes the action latent $u$ variable across all the time-steps. Both $z_t$s and $u_t$s are independent of across time.  
From the Causal relationship between $a_t$ and $x_t$ we get:
\begin{align}
\text{ln}q_{\zeta}(x_{1:T},a_{1:T})&\equiv
\text{ln}q_{\zeta}(x,a)= \ln q_{\zeta_2} (a|x)q_{\zeta_1}(x)\\
&= \ln q_{\zeta_2} (a|x)+ \ln q_{\zeta_1}(x) \label{eq:pf_causal_elbo0}
\end{align}
from \eqref{eq:pf_causal_elbo0}, we can derive the lower bound for $\ln q_{\zeta_1}(x) $ as:
\begin{align}
\text{ln}q_{\zeta_1}(x)&=\ln \int_{z} q_{\zeta_1} (x|z)p(z)\\
&=\ln \int_{z}q_{\zeta_1}(x|z)p(z)\frac{p_\theta(z|x)}{p_\theta(z|x)}\\
&=\ln \left(\E_{p_\theta(z|x)} q_{\zeta_1}(x|z)\frac{p(z)}{p_\theta(z|x)}\right)\\
&\geq \E_{p_\theta(z|x)} \left(\ln q_{\zeta_1}(x|z)\frac{p(z)}{p_\theta(z|x)}\right)\\
&=\E_{p_\theta(z|x)}\ln q_{\zeta_1}(x|z)-\E_{p_\theta(z|x)} \left(\ln\frac{p_\theta(z|x)}{p(z)}\right)\\
&=\E_{p_\theta(z|x)}\ln q_{\zeta_1}(x|z)- D_{KL}(p_\theta(z|x)||p(z))
\label{eq:pf_causal_elbo1}
\end{align}
Similar to VG-LeaP, we use RNN architectures  $\overset{\curvearrowleft}{RNN}_{\zeta_{1}}$ in Causal-LeAP to recursively predict image frames at each time-step $t$. $\overset{\curvearrowleft}{RNN}_{\zeta_{1}}$ takes the encoded past image $x_{t-1}$, action $a_{t-1}$ and $z_t$ as input. With the recursive behaviour of $\overset{\curvearrowleft}{RNN}_{\zeta_{1}}$, we approximate $\ln q_{\zeta_1}(x|z)\approx \ln q_{\zeta_1}(x_{1:T}|z_{1:T},a_{1:T-1})$ as:
\begin{align}
   \ln q_{\zeta_1}(x|z)\approx \ln q_{\zeta_1}(x_{1:T}|z_{1:T},a_{1:T-1})&=\ln\prod_{t}q_{\zeta_1}(x_{t}|x_{1:t-1},z_{1:T},a_{1:t-1},\cancel{a_{t:T-1}})\\
   &=\sum_t \ln q_{\zeta_1}(x_{t}|x_{1:t-1},z_{1:t},a_{1:t-1})
   \label{eq:pf_causal_elbox}
\end{align}
In the case of the posterior and learned prior of the image prediction networks $\overset{\curvearrowleft}{RNN}_{\theta}$ and $\overset{\curvearrowleft}{RNN}_{\phi}$ respectively in Causal-LeAP, we recursively feed the action image $x_t$ to approximate $z_t$ (in case of learned prior we feed $x_{t-1}$ ). Cause $z_t$s are independent across time,  $p_\theta(z|x)$ can be expressed as:
\begin{align}
   p_\theta(z_{1:T}|x_{1:T})&\equiv  p_\theta(z|x)=\prod_{t}p_\theta(z_t|x_t)
\end{align}
We can rewrite $D_{KL}(p_\theta(z|x)||p(z))$ as:
\begin{align}
   D_{KL}(p_\theta(z|x)||p(z))&\equiv D_{KL}(p_\theta(z_{1:T}|x_{1:T})||p(z_{1:T}))=\int_{z}p_\theta(z|x)\ln \frac{p_\theta(z|x)} {p(z)}\\
   &=\int_{z_1}\cdots \int_{z_T}p_\theta(z_1|x_1)\cdots  p_\theta(z_T|x_{1:T})\ln \frac{p_\theta(z_1|x_1)\cdots  p_\theta(z_T|x_{1:T})}{p(z_1)\cdots p(z_T)}\\
   &=\int_{z_1}\cdots \int_{z_T}p_\theta(z_1|x_1)\cdots  p_\theta(z_T|x_{1:T})\sum_t \ln \frac{p_\theta(z_t|x_{1:t})}{p(z_t)}\\
   &=\sum_t \int_{z_1}\cdots \int_{z_T}p_\theta(z_1|x_1)\cdots  p_\theta(z_T|x_{1:T}) \ln \frac{p_\theta(z_t|x_{1:t})}{p(z_t)}
\label{eq:pf_causal_kldx1}
\end{align}
Since $\int_z p_\theta(z)=1$ we can further simplify \eqref{eq:pf_causal_kldx1} as
\begin{align}
   D_{KL}(p_\theta(z|x)||p(z))&=\sum_t \int_{z_t}p_\theta(z_t|x_{1:t}) \ln \frac{p_\theta(z_t|x_{1:t})}{p(z_t)}\\
   &=\sum_t  D_{KL} (p_\theta(z_t|x_{1:t})||p(z_t))
\label{eq:pf_causal_kldx}
\end{align}
The lower bound of $\ln q_{\zeta_2} (a|x)$ is derived as follows:
\begin{align}
\text{ln}q_{\zeta_2}(a|x)&=\ln \int_{u} q_{\zeta_2} (a|u,x)p(u|x)\\
&=\ln \int_{u}q_{\zeta_2}(a|u,x)p(u|x)\frac{p_\psi(u|a,z)}{p_\psi(u|a,z)}\\
&=\ln \left(\E_{p_\psi(u|a,z)} q_{\zeta_2}(a|u,x)\frac{p(u|x)}{p_\psi(u|a,z)}\right)\\
&\geq \E_{p_\psi(u|a,z)} \left(\ln q_{\zeta_2}(a|u,x)\frac{p(u|x)}{p_\psi(u|a,z)}\right)\\
&=\E_{p_\psi(u|a,z)}\ln q_{\zeta_2}(a|u,x)-\E_{p_\psi(u|a,z)} \left(\ln\frac{p_\psi(u|a,z)}{p(u|x)}\right)\\
&=\E_{p_\psi(u|a,z)}\ln q_{\zeta_2}(a|u,x)- D_{KL}(p_\psi(u|a,z)||p(u|x))
\label{eq:pf_causal_elbo2}
\end{align}
Now combining \eqref{eq:pf_causal_elbo0}, \eqref{eq:pf_causal_elbo1} and \eqref{eq:pf_causal_elbo2} we get:
\begin{equation}
\begin{aligned}
\text{ln}q_{\zeta}(x,a)\geq\E_{p_\theta(z|x)}\ln q_{\zeta_1}(x|z)+\E_{p_\psi(u|a,z)}\ln q_{\zeta_2}(a|u,x)-& D_{KL}(p_\theta(z|x)||p(z))\\
&-D_{KL}(p_\psi(u|a,z)||p(u|x))
\label{eq:pf_causal_elbo}
\end{aligned}
\end{equation}
In the case of action predictor in Causal-LeAP, to predict $a_t$ we use  RNN architecture $\overset{\curvearrowleft}{RNN}_{\zeta_{2}}$ which takes the past action $a_{t-1}$ and $u_t$ as inputs at time $t$. Thus recursively it builds dependence upon all past actions $a_{1:t-1}$ and action latent variable $u_{1:t}$. Please note in the case of the action predictor network we do not feed the last image $x_t$ as input. We found that even without the image $x_t$ as input, the action predictor generates accurate approximation of future actions. Thus $\ln q_{\zeta_2}(a|u,x)\approx \ln q_{\zeta_2}(a|u) $ in  can be expressed as: 
\begin{align}
   \ln q_{\zeta_2}(a_{1:T}|u_{1:T})&\equiv \ln q_{\zeta_2}(a|u)=\ln\prod_{t}q_{\zeta_2}(a_{t}|a_{1:t-1},u_{1:T})\\
   &=\sum_t \ln q_{\zeta_2}(a_{t}|a_{1:t-1},u_{1:t})
   \label{eq:pf_causal_elboa}
\end{align}
In the case of the posterior and learned prior of the action prediction networks $\overset{\curvearrowleft}{RNN}_{\psi}$ and $\overset{\curvearrowleft}{RNN}_{\varphi}$ respectively in Causal-LeAP, we recursively feed the action $a_t$ and the image latent variable $z_t$ to approximate $u_t$ (in case of learned prior we feed $(a_{t-1},z_{t-1})$ ). Cause $u_t$s are independent across time,  $p_\theta(u|a,z)$ can be expressed as:
\begin{align}
   p_\psi(u_{1:T}|a_{1:T},z_{1:T})&\equiv  p_\psi(u|a,z)=\prod_{t}p_\psi(u_t|a_t,z_t)
\end{align}
We can rewrite $D_{KL}(p_\psi(u|a,z)||p(u|x))$ as:
\begin{align}
   D_{KL}&(p_\psi(u|a,z)||p(z))\equiv D_{KL}(p_\psi(u_{1:T}|a_{1:T},z_{1:T})||p(u_{1:T}|x_{1:T}))=\int_{u}p_\psi(u|a,z)\ln \frac{p_\psi(u|a,z)} {p(u|x)}\\
   &=\int_{u_1}\cdots \int_{u_T}p_\psi(u_1|a_1,z_1)\cdots  p_\psi(u_T|a_{1:T},z_{1:T})\ln \frac{p_\psi(u_1|a_1,z_1)\cdots  p_\psi(u_T|a_{1:T},z_{1:T})}{p(u_1|x_1)\cdots p(u_T|x_T)}\\
   &=\int_{u_1}\cdots \int_{u_T}p_\psi(u_1|a_1,z_1)\cdots  p_\psi(u_T|a_{1:T},z_{1:T})\sum_t \ln \frac{p_\psi(u_t|a_{1:t},z_{1:t})}{p(u_t|x_t)}\\
   &=\sum_t \int_{u_1}\cdots \int_{u_T}p_\psi(u_1|a_1,z_1)\cdots  p_\psi(u_T|a_{1:T},z_{1:T}) \ln \frac{p_\psi(u_t|a_{1:t},z_{1:t})}{p(u_t|x_t)}
\label{eq:pf_causal_klda1}
\end{align}
Since $\int_u p_\psi(u)=1$ we can further simplify \eqref{eq:pf_causal_klda1} as
\begin{align}
   D_{KL}(p_\psi(u|a,z)||p(u|x))&=\sum_t \int_{u_t}p_\psi(u_t|a_{1:t},z_{1:t}) \ln \frac{p_\psi(u_t|a_{1:t},z_{1:t})}{p(u_t|x_t)}\\
   &=\sum_t  D_{KL} (p_\psi(u_t|a_{1:t},z_{1:t})||p(u_t|x_t)
\label{eq:pf_causal_klda}
\end{align}
Now combining \eqref{eq:pf_causal_elbox}, \eqref{eq:pf_causal_elboa}, \eqref{eq:pf_causal_kldx} and \eqref{eq:pf_causal_klda} with \eqref{eq:pf_causal_elbo} we get the final expression for the variational lower bound as:
\begin{align}
    \text{ln}q_{\zeta}(x,a)&\geq\E_{p_\theta(z|x)}\ln q_{\zeta_1}(x|z)+\nonumber\\
    &\hspace{5em}+\E_{p_\psi(u|a,z)}\ln q_{\zeta_2}(a|u)- D_{KL}(p_\theta(z|x)||p(z))-D_{KL}(p_\psi(u|a,z)||p(u|x))\\
    &=\sum_t [\E_{p_\theta(z_{1:t}|x_{1:t})}\ln q_{\zeta_1}(x_{t}|x_{1:t-1},z_{1:t},a_{1:t-1})+\E_{p_\psi(u_{1:t}|a_{1:t},z_{1:t})}\ln q_{\zeta_2}(a_{t}|a_{1:t-1},u_{1:t}) \nonumber\\
    &\hspace{5em}-D_{KL} (p_\theta(z_t|x_{1:t})||p(z_t))-D_{KL} (p_\psi(u_t|a_{1:t},z_{1:t})||p(u_t|x_t)]
\label{eq:pf_causal_elbo_final}
\end{align}

\end{document}